\documentclass[sigconf]{acmart}
\setcopyright{none}
\renewcommand\footnotetextcopyrightpermission[1]{} 

\AtBeginDocument{%
  \providecommand\BibTeX{{%
    \normalfont B\kern-0.5em{\scshape i\kern-0.25em b}\kern-0.8em\TeX}}}

\begin{document}

\title{Drug Interaction Vectors Neural Network: DrIVeNN}
\subtitle{Modeling Polypharmacy Adverse Drug Events and a Case Study in Cardiovascular Disease Treatment}

\author{Natalie Wang}
\email{nwang42@jhu.edu}
\affiliation{%
  \institution{Johns Hopkins University}
  \city{Baltimore}
  \state{MD}
  \country{USA}
}

\author{Casey Overby Taylor}
\email{cot@jhu.edu}
\affiliation{%
  \institution{Johns Hokins University}
  \city{Baltimore}
  \state{MD}
  \country{USA}}

\begin{abstract}
  Polypharmacy, the concurrent use of multiple drugs to treat a single condition, is common in patients managing multiple or complex conditions. However, as more drugs are added to the treatment plan, the risk of adverse drug events (ADEs) rises rapidly. Many serious ADEs associated with polypharmacy only become known after the drugs are already in use. It is impractical to test every possible drug combination during clinical trials. This issue is particularly prevalent among older adults with cardiovascular disease (CVD) where polypharmacy and ADEs are commonly observed. In this research, our primary objective was to identify key drug features and build and evaluate a model for modeling polypharmacy ADEs. Our secondary objective was to assess our model on a domain-specific case study. 
  
  We developed a two-layer neural network that incorporated drug features such as molecular structure, drug-protein interactions, and mono drug side effects (DrIVeNN). We assessed DrIVeNN using publicly available side effect databases and determined Principal Component Analysis (PCA) with a variance threshold of 0.95 as the most effective feature selection method. DrIVeNN performed moderately better than state-of-the-art models like RESCAL, DEDICOM, DeepWalk, Decagon, DeepDDI, KGDDI, and KGNN in terms of AUROC for the drug-drug interaction prediction task. 
  
  We also conducted a domain-specific case study centered on the treatment of cardiovascular disease (CVD). When the best performing model architecture was applied to the CVD treatment cohort, there was a significant increase in performance from the general model. We observed an average AUROC for CVD drug pair prediction increasing from 0.826 (general model) to 0.975 (CVD specific model). Our findings indicate the strong potential of domain-specific models for improving the accuracy of drug-drug interaction predictions. In conclusion, this research contributes to the advancement of predictive modeling techniques for polypharmacy ADEs. 
\end{abstract}



\settopmatter{printfolios=true}
\maketitle
\pagestyle{plain}
\section{Introduction}

Polypharmacy, characterized by the concurrent use of multiple drugs to treat a single condition, is a common in patients managing complex and terminal diseases. However, as more drugs are added to the treatment plan, the risk of adverse drug events (ADEs) increases rapidly.\cite{offsides_and_twosides} These ADEs can have severe consequences on patient health and well-being. Additionally, the prevalence of polypharmacy has been steadily increasing over the years, reaching a rate of 37\% in 2022 compared to 8.2\% in 1999-2000 and 15\% in 2011-2012.\cite{Kantor2015}\cite{Delara2022} In clinical studies, it is infeasible to test every possible combination of drugs for ADEs, so serious polypharmacy-related ADEs may only become known once the drugs  already in use, leading to a substantial healthcare burden. In the United States alone, these ADEs contribute to an estimated \$62 billion and nearly 150,000 premature deaths.\cite{Garber2019} As the number of available drugs continues to escalate, traditional experimental methods for assessing drug-drug interactions (DDIs) struggle to keep pace, necessitating innovative computational approaches to accelerate our understanding.   
 
We sought to build a novel deep learning model for DDI prediction. Numerous studies show the potential of deep learning models for the DDI prediction task. These include RESCAL, DEDICOM, DeepWalk, Decagon, DeepDDI, KGDDI, and KGNN\cite{rescal, dedicom, deepwalk,Zitnik2018, deepddi, kgddi, kgnn}. Through a review of the literature, we extracted some key features and takeaways from each model, which served as a basis of comparison for our model. 

Prior models use deep learning as a tool to discover hidden interactions between drug features in many different ways. For example, RESCAL uses a latent factorization method to represent drugs and interactions as low-dimensional vectors. Another model, DEDICOM, uses tensor decomposition to uncover latent features in drug-drug interaction data. Additionally, DeepWalk uses a graph embedding technique to capture structural information of drug interaction networks.

Other notable models include Decagon, a tensor factorization decoder, and DeepDDI which incorporated domain knowledge via drug chemical structures and protein sequence information. KGDDI and KGNN both exploit knowledge graphs to capture semantic relationships between drugs, proteins, and diseases. These two models use graph neural networks to capture complex interactions. In our research, we utilized these state-of-the-art models as baselines for comparison with our proposed model. 

Additionally, we built and evaluated a disease-specific model. We used cardiovasular disease (CVD) as a case study for other conditions that often require polypharmacy. Older adults with CVD are particularly prone to polypharmacy. A retrospective chart review at a tertiary care center found among older adults (65 years and above) with a history of CVD and admitted to the cardiology service, the prevalence of polypharmacy was 95\%, hyper-polypharmacy 69\%, and at least one severe potential drug-drug interaction (DDI) 77.5\%.\cite{Marwan} To the best of our knowledge, a domain-specific model has not yet been investigated for the drug-drug interaction task. 

Our overall approach expands on existing work by integrating diverse sources of drug-related data and a new model structure. The Drug Interaction Vectors Neural Network (DrIVeNN) model we build uses graph neural networks to learn complex patterns from datasets consisting of known drug interactions, drug targets, and molecular structures. We also drew inspiration from a study conducted by Chen et al (2021), who developed a deep learning method called MUFFIN that incorporates chemical structures, drug-target interactions, and known drug side effects to predict DDIs.\cite{Chen2021} Their results demonstrate the effectiveness of multi-modal data integration for the DDI prediction task and motivated us to include these same data types for our model. Lastly, DrIVeNN incorporates an ensemble method for prediction, inspired by the works of Masumashah et al (2021).\cite{Masumshah2021}  

\section{Objectives}
\begin{itemize}
    \item Build and evaluate the DrIVeNN model with diverse drug feature vectors for the drug-drug interaction prediction task.
    \item Evaluate DrIVeNN for the DDI prediction task in the CVD domain, as a case study for domain-specific DDI prediction tasks.
\end{itemize}
\section{Methods}
\subsection{Datasets}
To address our first objective, we employed the datasets compiled and preprocessed by Zitnik et al\cite{Zitnik2018} which contained drug-drug interactions, drug-protein interactions, and mono drug side effects. The authors used the STITCH (Search Tool for InTeractions of CHemicals) database as a primary source for drug-protein interactions. STITCH is a comprehensive database of protein-chemical interaction data with original data sources for each interaction.\cite{stitch} To capture mono drug side effects, Zitnik et al\cite{Zitnik2018} used two distinct databases: SIDER (Side Effect Resource) and OFFSIDES. SIDER contains drug-side effect associations obtained from drug label text, while OFFSIDES contains drug-side effect associations generated from adverse event reporting systems.\cite{sider, offsides_and_twosides} To obtain the structural attributes of drugs, we used the SMILES (Simplified Molecular Input Line Entry System) representation of each drug to construct a molecular graph and employed a graph neural network to generate structural representations. 

To address our second objective, we utilized the Inxight Drugs API to conduct a comprehensive search for drugs prescribed for three major cardiovascular diseases: myocardial infarction (MI), congestive heart failure (CHF), and coronary artery disease (CAD).\cite{ncats} Through this analysis, we found 30 distinct principal forms of drugs for MI, 22 for CHF, and 18 for CAD. To integrate this data with our existing datasets, we gathered Unique Ingredient Identifiers (UNIIs) and InCHIKeys for each cardiovascular disease treatment drug. We chose these because they were the most frequently occurring in the dataset. For drugs in our other drug datasets, which were identified solely by PubChem IDs, we used a two-step process to obtain their corresponding UNIIs. First, we downloaded UNII drug records from open.fda.gov and matched records that contained both UNII and PubChem IDs, thus obtaining UNIIs for the drugs in our dataset. However, approximately 300 drugs from our original dataset still lacked UNII matches so we manually used the PubChem Lookup tool to find to find their drug names and InCHIKeys, then the Global Substance Registration System (GSRS) to search for their UNIIs based on their drug names and InCHIKeys.\cite{pubchem, gsrs} There is an overview of this process shown in Figure 1. The CVD treatment drugs we identified that were also present in our DDI datasets were Enoxaparin Sodium, Niacin, Aspirin, Carvediolol, Metoprolol, Nitroglycerin, Ramipril, Valsartan, Amlodipine, Ticlopidine, Chlorothizide, Ethacrynic Acid, Indapamide, Metolazone, Ramipril, and Spironolactone.
\begin{figure*}[h]
  \centering
  \includegraphics[width=\linewidth]{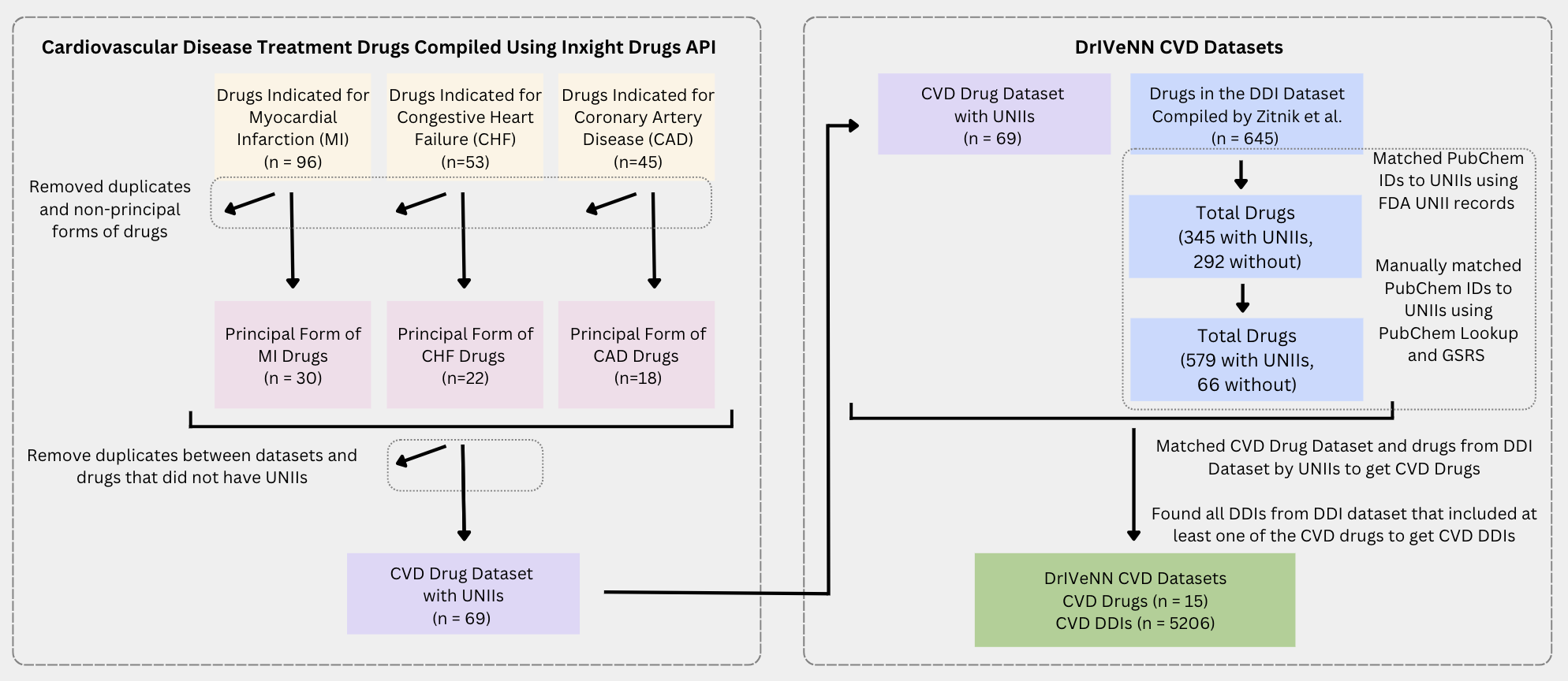}
  \caption{DrIVeNN Cardiovascular Disease Data Pipeline.}
  \Description{Figure 1 summarizes the CVD data pipeline. To gather a list of CVD treatment drugs, we performed the following steps: 1. Used the Inxight Drugs API to find drugs indicated for MI, CHF, and CAD, 2. Removed duplicates and non-principal forms of the drugs, and 3. Selected unique drugs within the three sets that contained UNIIs, this became our CVD treatment drug dataset. The steps in the second part of the pipeline were: 1. Matched PubChem IDs in DDI dataset with UNIIs using UNII records from fda.gov, 2. Manually search unmatched drugs from the DDI dataset for matching UNIIs using PubChem Lookup and GSRS, and 3. Matched CVD drugs found in the first part of the pipeline and our DDI drugs based on UNIIs then found all DDIs that included at least one CVD drug to create our CVD DDI dataset.}
\end{figure*}

\subsection{Data-Driven Motivation for this Approach}
During our exploratory data analysis, we made several observations that served as our motivation for creating a domain-specific model.

First, we noticed a difference between the ten most common mono side effects of all drugs and the ten most common mono side effects of drugs identified as CVD treatment drugs. Only two side effects, "emotional distress" and "blood creatinine increased" overlapped both groups. In the context of this study, a mono side effect refers to an ADE that is known to be caused by a drug (as opposed to a drug-drug interaction which is an ADE known to be caused by two drugs). This observation suggests potential differences in the drug features of CVD treatment drugs compared to other drugs, indicating the need for a specialized model for this domain.

We also observed the median number of drug-drug interactions (DDIs) for different drug sets. The median number of DDIs per all drug pairs was found to be 53. In contrast, drug pairs that contained at least one CVD treatment drug exhibited a higher median number of DDIs, 65 interactions, and the drug pairs the consisted of two CVD drugs, reached a median of 124 interactions. This notable increase in the likelihood of ADEs for CVD treatment drugs further motivated us to explore this domain.

\subsection{Data Processing}
Each drug was associated with three distinct features: drug structure features, drug-protein interaction features, and mono drug side effect features. We created two drug feature datasets, one with drug structure features and one without. To extract the drug structure features, we utilized dgllife, a Python package designed for deep learning on graphs.\cite{dgllife} We used a message-passing graph neural network, MPNN, that was pretrained for molecules to extract corresponding drug structure embeddings from our drug dataset. For the drug-protein interaction features and mono drug side effect features, we applied Principle Component Analysis (PCA) as a feature extraction method\cite{scikit}. We also applied normalization techniques and UMAP (Uniform Manifold Approximation and Projection) to address sparsity in the data\cite{scikit}. To capture the full representation of each drug, we concatenated the three types of features into a single drug feature vector (Fig. 2). To represent a drug-drug pair, we summed the individual drug features following protocols from similar studies (Fig. 3a)\cite{Masumshah2021}. 

\begin{figure*}[h]
  \centering
  \includegraphics[width=\linewidth]{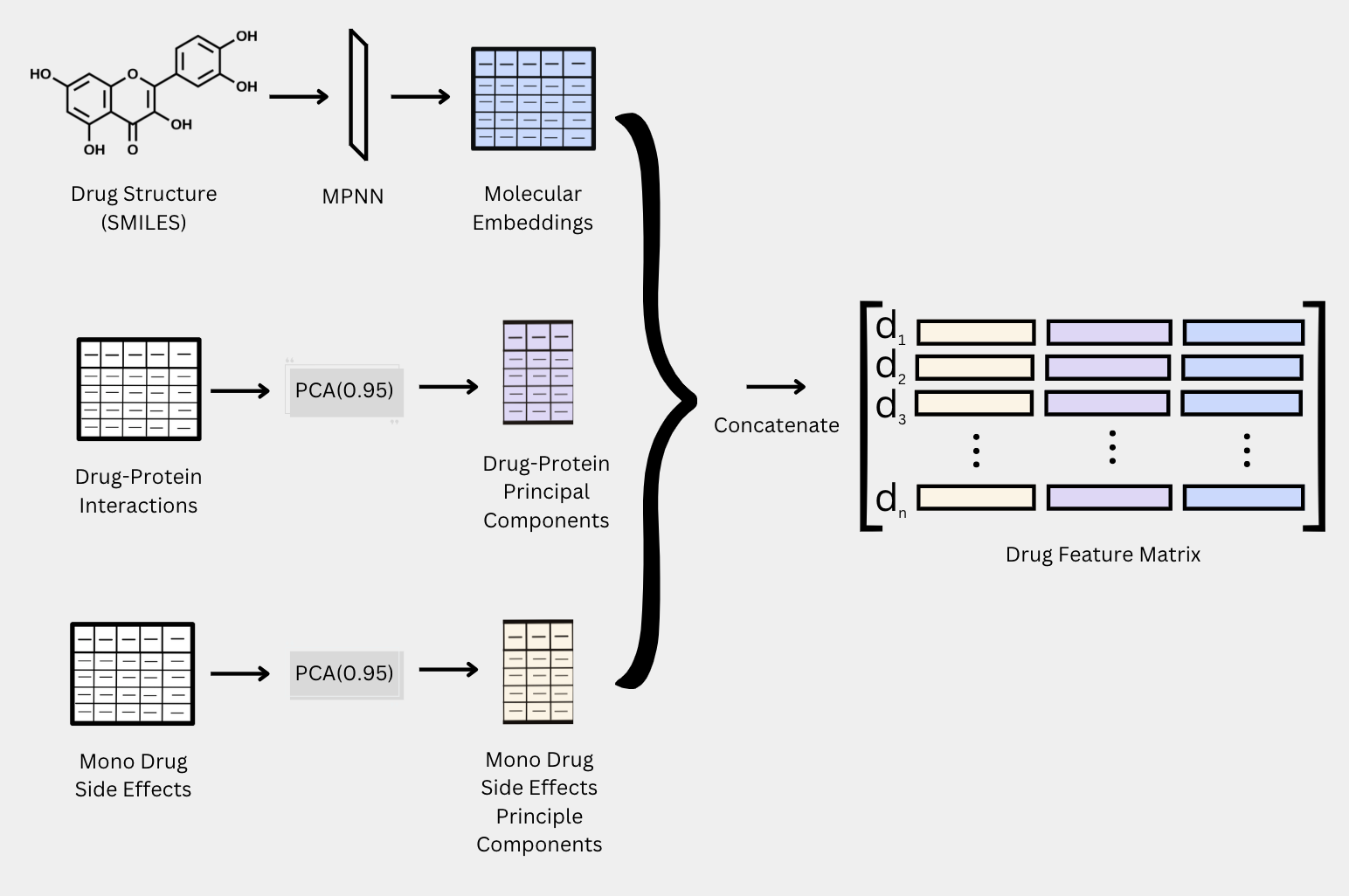}
  \caption{Drug feature matrix creation for drugs $d_1, d_2, ..., d_n$. This summarizes the drug feature matrix creation process which includes: 1. Applying MPNN, a graph neural network, to the drug structure dataset, 2. Training and applying PCA(0.95) to the drug-protein dataset and mono drug side effect dataset, and 3. Concatenating the three separate processed datasets into the drug feature matrix where each row represents one drug.}
  \Description{Figure 2 summarizes the drug feature matrix creation process which includes: 1. Applying MPNN, a graph neural network, to the drug structure dataset, 2. Training and applying PCA(0.95) to the drug-protein dataset and mono drug side effect dataset, and 3. Concatenating the three separate processed datasets into the drug feature matrix where each row represents one drug.}
\end{figure*}

\begin{figure*}[h]
  \centering
  \includegraphics[width=\linewidth]{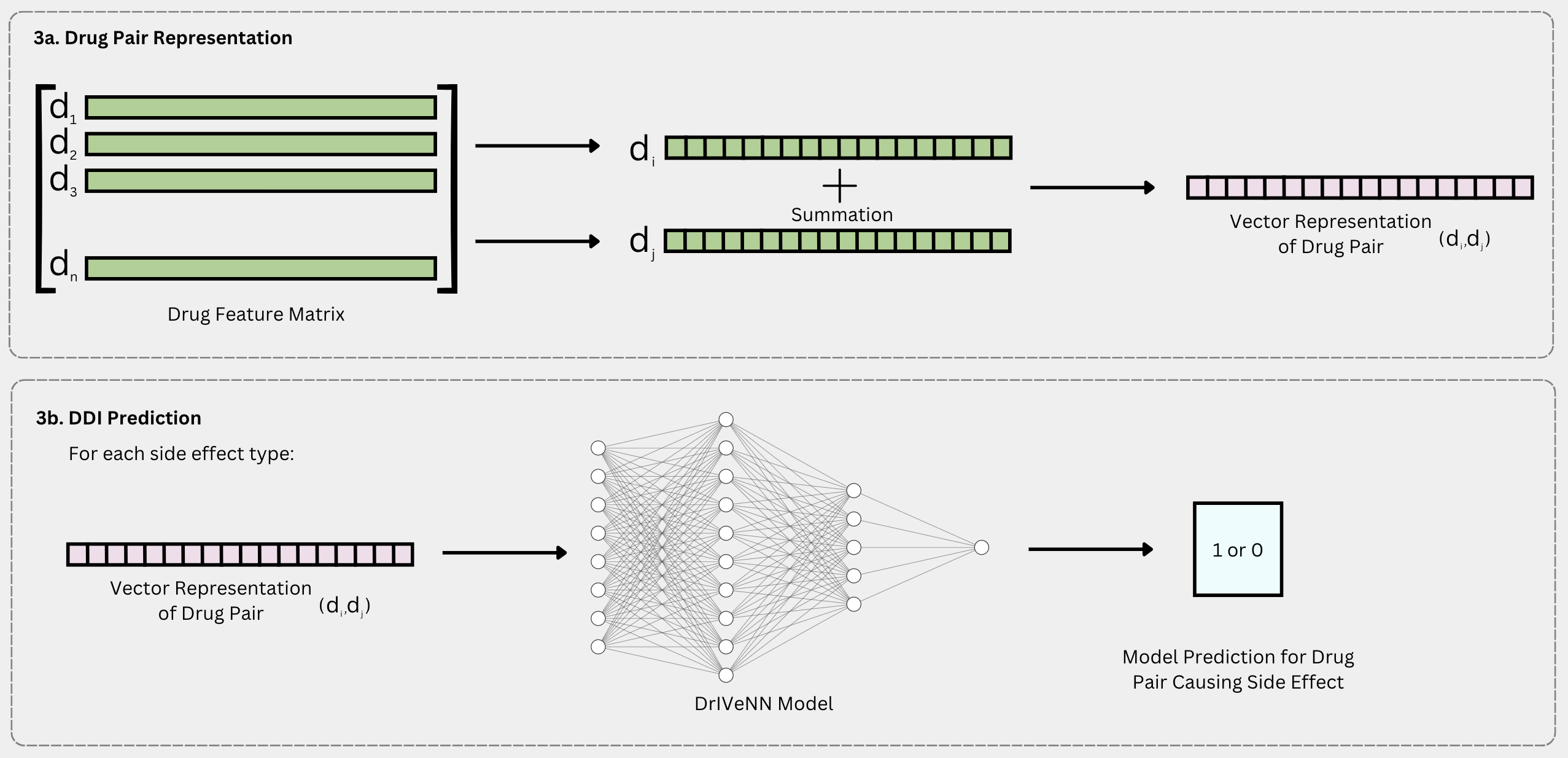}
  \caption{Drug Pair Prediction: 3a. Drug pair representation for drugs $(d_i, d_j)$. Element-wise summation was performed on the individual drug representations to create one representative vector. This process was done for all drug pairs in the dataset. 3b. DDI prediction process for a single drug pair and for a single side effect}
  \Description{This graphic shows the drug pair representation and prediction process.}
\end{figure*}

To construct our classification dataset, we implemented negative sampling following the established protocol from other studies.\cite{Zitnik2018, Trouillon2016, Mikolov2013} For each side effect, there are corresponding drug-drug pairs that were identified as causative factors, these are our positive edges.  To generate negative edges, we randomly selected drug pairs that were not associated with the given side effect. This approach allowed us to create a balanced dataset that included both positive and negative instances, helping with the classification task and ensuring more complete coverage of potential drug-drug interactions. 
\subsection{Model Training}
For every type of side effect, we partitioned the corresponding drug-drug pairs into train, validation, and test sets using an 80/10/10 ratio, respectively. This partitioning ensured the model was trained on a sufficiently large dataset while also providing separate datasets for validation and final evaluation. During the training phase, our objective was to minimize the binary cross-entropy loss function. Each training run consisted of 50 epochs, allowing the model to learn and adjust its parameters over multiple iterations. We used a pretrained GNN for the data extraction step, so our model training consists of learning weights for our predictive feed-forward neural network. 

\subsection{Experiments}
In our initial experiment, we examined different feature selection methods to address sparsity observed in the dataset and improve computational efficiency. We evaluated various levels of Principle Component Analysis (PCA) and a combination of normalization and Uniform Manifold Approximation and Projection (UMAP). PCA was chosen as a dimensionality reduction technique to retain the most informative components of the drug features, while mitigating the effect of sparse data on our analysis. We utilized normalization and UMAP as complementary approaches. Because our drug feature vectors were derived from multiple sources, we first applied normalization to ensure features were on the same scale for fair comparisons. We chose UMAP due to its well-known effectiveness at preserving both local and global structure within the data while reducing its dimensionality.\cite{umap} For this experiment, we used the model architecture developed by Masumashah et al, a three-layer neural network with 300, 200, and 100 nodes for each respective layer\cite{Masumshah2021}.  This architecture has demonstrated promising performance in previous studies and served as a foundation for our experiments. One key difference between that study and ours is the incorporation of molecular drug data into the model. 

In the second experiment, we performed hyperparameter tuning on the model using the Hyperband algorithm which is a competition-based approach to efficiently explore hyperparameter space.\cite{hyperband} We considered variables such as number of layers, neurons per layer, the inclusion of batch normalization, and dropout regularization. Dropout was considered to prevent overfitting and promote generalization while batch normalization was incorporated to normalize inputs within each batch which can lead to faster convergence. 

For our third experiment, we conducted training using our best-performing feature selection methods and model architecture on our cardiovascular disease treatment dataset. This dataset is comprised of drug pairs where at least one drug was specifically used for cardiovascular disease treatment. By focusing on this domain-specific dataset, we aimed to assess the performance and applicability of our approach within a specific medical context. 

\section{Evaluation and Performance}
For each model examined in our study, we evaluated its performance using the following metrics: AUROC (Area Under ROC Curve) and AUPR (Area Under the Precision-Recall Curve). We used AUROC to provide an assessment of the model’s discriminatory power and AUPR to capture the trade-off between precision and recall. We believe AUROC is a robust evaluation metric because our test set is balanced. As mentioned in the data processing step, we employed negative sampling techniques to get a balanced dataset. We calculated the average of these metrics across the different side effects to provide a more comprehensive evaluation of the model’s performance. We also evaluated each model on training time.

Additionally, we investigated side effects for which the model performed exceptionally well and those where its performance was comparatively poorer. To further understand the significance of these side effects, we employed Saedr scores as a means of assessing the severity of side effects grouped by model prediction accuracy. Saedr scores, developed by Lavertu et al, are designed to quantify the severity of adverse drug events using social media network analysis.\cite{Lavertu2021} We categorized side effects into three distinct bins based on the model prediction accuracy: AUROC 0.85-0.90, AUROC 0.90-0.95, and AUROC 0.95-0.99. Then we evaluated and compared the average Saedr scores associated with each of these bins.

\section{Results}
\subsection{Feature Selection}
The initial dimensions of our drug feature matrix, before any feature selection, were (645, 18279) across 964 side effects. During our first experiment, we assessed the impact of different feature selection methods on the performance of our model (Table 1). We observed as the variance captured by PCA increased and as molecular embeddings were incorporated into the input matrix, there was a moderate improvement in AUROC and AUPRC. We chose to proceed with PCA 0.95 as the chosen feature selection method and with molecular embeddings for our subsequent experiments. In the future, it may also be interesting to look more closely at the performance of some of the PCA datasets with lower variance as they would utilize less computational resources and time.
\begin{table*}
  \caption{Evaluation of Feature Selection Methods}
  \label{tab:freq}
  \begin{tabular}{cccccl}
    \toprule
    &&With Molecular Embeddings&&Without Molecular Embeddings&\\
    \toprule
    Feature Selection&Drug Feature Dimensions&AUROC&AUPRC&AUROC&AUPRC\\
    \midrule
    \ UMAP + Norm & $\left( 645, 2 \right)$ & 0.516 & 0.522 & 0.532 & 0.545 \\
    \ PCA(0.85) & $\left( 645, 667 \right)$ & 0.901 & 0.824 & 0.897 & 0.819 \\
    \ PCA(0.90) & $\left( 645, 733 \right)$ & 0.900 & 0.823 & 0.901 & 0.821 \\
    \ PCA(0.95) & $\left( 645, 825 \right)$ & 0.903 & 0.826 & 0.902 & 0.825\\
    \ PCA(0.99) & $\left( 645, 972 \right)$ & 0.906 & 0.829 & 0.905 & 0.825\\
  \bottomrule
\end{tabular}
\end{table*}

\subsection{Hyperparameter Tuning}
Upon completion of the hyperparameter turning process, we analyzed the results obtained for each side effect. To ensure a more generalizable mode, we selected the overall best model parameters by identifying the most common parameters across all side effects.  This approach aimed to find a balance between individual side effect performance and the ability to capture patterns across collective characteristics. The selected hyperparameters which reflect our overall best model configuration are highlighted in Table 2. 

\begin{table}
  \caption{Selected Hyperparameter Values}
  \label{tab:freq}
  \begin{tabular}{cccl}
    \toprule
    Number of Layers&Neurons Per Layer&Batch Norm&Dropout\\
    \midrule
    \ 2 & 300, 100 & Yes & No \\
  \bottomrule
\end{tabular}
\end{table}

\subsection{Overall Model Performance}
In our analysis, we found that our hyperparameter-tuned model, DrIVeNN, slightly outperformed some commonly used baselines for AUROC (Table 3). Please note that the performance values for AUROC and AUPRC for other models were obtained from previously published works.\cite{Masumshah2021} On CPU, the DrIVeNN model took about 2 hours and 30 minutes to train. 
\begin{table}
  \caption{Evaluation of General Model Performance}
  \label{tab:freq}
  \begin{tabular}{ccl}
    \toprule
    Model&AUROC&AUPRC\\
    \midrule
    \ RESCAL & 0.693 & 0.613 \\
    \ DEDICOM & 0.705 & 0.637 \\
    \ DeepWalk & 0.761 & 0.737 \\
    \ DeepDDI & 0.830 & 0.503 \\
    \ Decagon & 0.874 & 0.825 \\
    \ KGDDI & 0.891 & 0.653 \\
    \ KGNN & 0.896 & 0.658 \\
    \ DrIVeNN\_all & \textbf{0.901} & 0.821 \\
  \bottomrule
\end{tabular}
\end{table}

\subsection{Domain-Specific Results}
The performance evaluation of our domain-specific model, specifically focused on cardiovascular disease treatment drug pairs, yielded promising results. We observed a significant increase in AUROC and AUPRC values, reaching 0.9725 and 0.952 respectively (Table 4). For comparison, we evaluated the performance of the general DrIVeNN\_all model on the same datasets. The general model had an average AUROC of 0.901 for the test set created from all drug pairs, and only 0.701 for the test set created from CVD drug pairs. These findings show our domain-specific model performs significantly better than a general model on domain-specific drug pairs. These findings highlight the potential of domain-specific models to increase predictive performance for drug-drug interaction prediction. Additionally, on CPU, our domain-specific model trained in under 20 minutes. 

\begin{table}
  \caption{Evaluation of Domain-Specific Models}
  \label{tab:freq}
  \begin{tabular}{cccl}
    \toprule
    Model&Test Set&AUROC&AUPRC\\
    \midrule
    \ DrIVeNN\_all & All Drug Pairs & 0.901 & 0.821 \\
    \ DrIVeNN\_all & CVD Drug Pairs & 0.826 & 0.802 \\
    \ DrIVeNN\_cvd & All Drug Pairs & 0.701 & 0.683 \\
    \ DrIVeNN\_cvd & CVD Drug Pairs &\textbf{0.975} & \textbf{0.952} \\
  \bottomrule
\end{tabular}
\end{table}

\subsection{Side Effect Severity}
We conducted an analysis of side effects associated with CVD-related treatment polypharmacy, specifically focusing on the prediction AUROC of these side effects from our model, DrIVeNN\_cvd. To address the severity of these side effects, we utilized Saedr scores as a metric. Out of the 3,196 unique side effects caused by CVD-related treatment polypharmacy, we were able to obtain Saedr scores for 2,794 side effects. This subset of side effects became the focus of our analysis.  In the future, we would like to explore additional severity metrics and work on obtaining a larger domain-specific dataset to allow for a more comprehensive analysis of drug-drug interactions within the domain. 

After excluding the first group (<0.85 AUROC) from our analysis because it contained only one side effect, we observed that, on average, DrIVeNN\_cvd demonstrated slightly higher AUROC scores for side effects with higher severity scores (Table 5). This finding suggests that our model exhibits improved performance on side effects that are associated with greater severity, as measured by Saedr scores. To provide illustrative examples, we have included a selection of side effects from each group in the table below to showcase the variation in severity and corresponding AUROC scores. 
\begin{table*}
  \caption{Severity of Side Effects by AUROC Bin}
  \label{tab:freq}
  \begin{tabular}{cccl}
    \toprule
    AUROC Bin&Median Saedr Score&Selected Side Effects in Group&Total Side Effects\\
    \midrule
    \ $\left( 0.85, 0.90 \right)$& 0.525& carpal tunnel, hematoma, constipation, glaucoma & 8\\
    \ $\left(0.90, 0.95\right)$ & 0.588& cardiac murmur, angina, excess potassium, incontinence  & 32\\
    \ $\left(0.95, 0.99\right)$ & 0.605& cardiac ischemia, heart attack, phlebothrombosis, ventricular fibrillation  & 239\\
  \bottomrule
\end{tabular}
\end{table*}

\section{Discussion}
The drug-drug interaction (DDI) prediction task is a critical area of research that plays a role in ensuring patient safety and healthcare management. Traditionally, DDI prediction has been approached in two distinct ways: predicting whether a side effect will occur as a result of drug interactions and predicting the specific side effect that may occur. Our study focuses on the second aspect. By targeting the prediction of specific side effects, we have a more granular approach to understanding the potential outcomes of polypharmacy. In our evaluation, we introduced Saedr scores as a metric to assess the severity of side effects with lower and higher prediction accuracy to gain insights into the potential impact and clinical relevance of this model. This approach allows us to identify where areas of improvement in predication accuracy may have significant clinical implications. 

This study presented DrIVeNN, a novel approach to DDI prediction that incorporated diverse drug features. In our methodology, we used PCA with a variance threshold of 0.95 for feature selection then for each drug pair, a representation was created by summing the selected features. This served as the input for our model which had a performance comparable to state-of-the-art baselines. Additionally, our findings indicate that our domain-specific model, DrIVeNN\_cvd, outperformed our general model in DDI prediction. This indicates the potential for improved accuracy to predict DDIs with domain-specific models. To the best of our knowledge, DrIVeNN\_cvd represents the first domain-specific model developed for the DDI prediction task. Exploring additional domains and developing tailored models could lead to further improvements in predictive accuracy. 

The first notable limitation is the size of our domain-specific dataset. Although we made substantial effort to collect relevant data for cardiovascular disease treatment, the dataset’s size may restrict the full exploration of the DDIs in the domain. Future work may include compiling a more comprehensive and expansive domain-specific dataset. In addition, it is worth exploring other medical domains where polypharmacy is common, for example, the mental health domain. This could offer insights into the generalizability and effectiveness of domain-specific models. Another limitation of our study is that we evaluated our findings solely on the dataset of known side effects. While this dataset provides a foundation for initial model evaluation, it does not fully encompass the complexities of real-world patient data. To better understand the potential to use this model to predict DDIs in a real-world clinical setting, we believe that evaluations on patient data sources, such as electronic health records (EHRs) are needed. Doing so would provide more insight into the applicability of the model in clinical settings and an evaluation framework for other models.  

Some other potential avenues for future research includes a more comprehensive exploration of both the time and space complexities of the model. Additionally, trying local Principal Component Analysis (PCA) could prove beneficial. While global PCA was employed due to its widespread application in contexts such as ours, it is worth noting its use comes with strong linearity assumptions. Consequently, global PCA might offer a better fit for representing the data.

The potential impact of DrIVeNN extends beyond drug development and clinical decision-making. Accurate DDI prediction can assist healthcare professionals in proactively mitigating adverse effects, optimizing drug combinations, and maximizing therapeutic outcomes. Additionally, pharmaceutical companies may benefit from using our model to optimize drug discovery pipelines, prioritize candidate drugs with lower interaction risks, and minimize costly experimental testing. 

The main contributions of this research paper can be summarized as follows: 
\begin{itemize}
    \item Novel Drug Feature Vectors and Model: Different from baseline models, we used drug feature vectors including drug structures, drug-protein interactions, and mono drug side effects. We also propose a feature selection method and model architecture for the drug-drug interaction task.
    \item Domain-Specific Model Performance: We evaluated our model in two ways: with all available drug pairs (as others have), and second, with a focus on polypharmacy in CVD treatment (new in this study). Experimental results obtained for this evaluation are noteworthy as they highlight the promising potential of the effectiveness of domain-specific models in DDI prediction.  
\end{itemize}
 
\section{Conclusion}
In conclusion, our research has highlighted the potential of domain-specific models for drug-drug interaction (DDI) prediction. By introducing DrIVeNN as a new model that utilizes diverse drug features and feature selection techniques, we have demonstrated competitive performance for the DDI prediction task, comparable to existing state-of-the-art models in terms of AUROC and AUPRC. We also have preliminary results suggesting our domain-specific model specifically tailored for cardiovascular disease treatment (DrIVeNN\_cvd), exhibits potential advantages in predictive accuracy. Furthermore, we observed that on average, DrIVeNN\_cvd performs better on more severe side effects, as quantified by the Saedr score system. This finding may suggest other underlying factors related to severity that influence predictive accuracy. Exploring these factors may provide more insights into the nature of drug-drug interactions. 

\bibliographystyle{ACM-Reference-Format}
\bibliography{export}



\end{document}